\documentclass{article}
\usepackage{ijcai26}

\usepackage{times}
\usepackage{soul}
\usepackage{url}
\usepackage[hidelinks]{hyperref}
\usepackage[utf8]{inputenc}
\usepackage[small]{caption}
\usepackage{graphicx}
\usepackage{amsmath}
\usepackage{amsthm}
\usepackage{booktabs}
\usepackage{algorithm}
\usepackage{algorithmic}
\usepackage[switch]{lineno}

\usepackage{amsmath,amsfonts,bm}

\def\eqref#1{equation~\ref{#1}}

\def\1{\bm{1}}

\def\vx{{\bm{x}}}

\def\vz{{\bm{z}}}

\def\mF{{\bm{F}}}

\def\mI{{\bm{I}}}

\def\mK{{\bm{K}}}

\def\mM{{\bm{M}}}

\def\mQ{{\bm{Q}}}

\def\mS{{\bm{S}}}

\def\mV{{\bm{V}}}
\def\mW{{\bm{W}}}
\def\mX{{\bm{X}}}

\def\mZ{{\bm{Z}}}

\DeclareMathAlphabet{\mathsfit}{\encodingdefault}{\sfdefault}{m}{sl}
\SetMathAlphabet{\mathsfit}{bold}{\encodingdefault}{\sfdefault}{bx}{n}

\def\sR{{\mathbb{R}}}

\usepackage{xspace}
\makeatletter
\DeclareRobustCommand\onedot{\futurelet\@let@token\@onedot}
\def\@onedot{\ifx\@let@token.\else.\null\fi\xspace}

\def\eg{\emph{e.g}\onedot}

\def\etal{\emph{et al}\onedot}
\makeatother

\usepackage[table]{xcolor}
\usepackage{multirow}
\usepackage{float}
\usepackage{colortbl}

\usepackage{pifont} 

\usepackage{natbib}

\makeatletter
\newcommand{\thickhline}{%
\noalign {\ifnum 0=`}\fi \hrule height 1pt
\futurelet \reserved@a \@xhline
}
\makeatother

\title{TextGaze: Prompting Gaze Target Estimation with Textual Scene Cues}

\author{
Junhui She$^{1,3}$
\and
Fei Wang$^{2,3,}$\thanks{Corresponding authors}
\and
Kun Li$^{5}$
\and
Yiqi Nie$^{3,4}$
\and
Yuxin Liu$^{3,4}$
\and \\
Zhangling Duan$^{3}$
\And
Xun Yang$^{1,*}$\\
\affiliations
$^1$University of Science and Technology of China \\
$^2$Hefei University of Technology 
\\
$^3$Institute of Artificial Intelligence, Hefei Comprehensive National Science Center\\
$^4$ Anhui University \\
$^5$ United Arab Emirates University \\
\emails
\{shejunhui323, ifei17.hfut, kunli.hfut, nieyiqi5, yuxinliu221\}@gmail.com,\\
duanzl1024@iai.ustc.edu.cn,
xyang21@ustc.edu.cn
}

\begin{document}

\maketitle

\begin{abstract}
Gaze target estimation aims to infer the position of a person's gaze within a scene. Within mainstream design logic, multi-branch methods require extra supervision and annotations, while streamlined designs prioritize low-level visual saliency over true gaze intent. The former leads to a high annotation burden and hinders domain transfer, whereas the latter causes misalignment between predicted attention and actual gaze targets. 
To address this issue, we propose TextGaze, a unified cross-modal architecture that leverages a Large Vision-Language Model (LVLM) as scalable semantic guidance to balance the two design paradigms. The model extracts visual features from a frozen encoder and utilizes an LVLM to obtain gaze-aligned textual cues. We design a transformer-based fusion module with hierarchical text supervision to preserve task semantics. Lightweight decoding heads enable the joint prediction of gaze heatmaps and in-/out-of-frame status. 
We evaluate our method on four mainstream datasets, and the results show competitive performance across key metrics with robust cross-dataset generalisation without extra fine-tuning. Overall, we provide a streamlined alternative to traditional designs and highlight the potential of LVLMs as accessible auxiliary guidance for gaze estimation. All contents are available at: \url{https://github.com/idremo/TextGaze-IJCAI2026}.
\end{abstract}

\section{Introduction}\label{introduction}
Gaze target estimation~\cite{recasens2017following,chong2018connecting,miao2023patch,ryan2025gaze,liu2024depth} is a task in computer vision that aims to predict a person’s gaze direction in a scene or determine if the gaze target lies within the image. As human gaze is a fundamental non-verbal cue reflecting attention allocation, intent, and cognitive state~\cite{eckstein2017beyond}, it is essential for human behavior understanding~\cite{li2023data,li2025prototypical,li2026ma,wang2024frequency}, with broad applications in human-computer interaction~\cite{katsini2020role,wang2026xinsight,chen2025towards}, extended reality~\cite{sitzmann2018saliency}, and education~\cite{wang2025robust}.

\begin{figure}[t!]
\centering
\centerline{\includegraphics[width=\columnwidth]{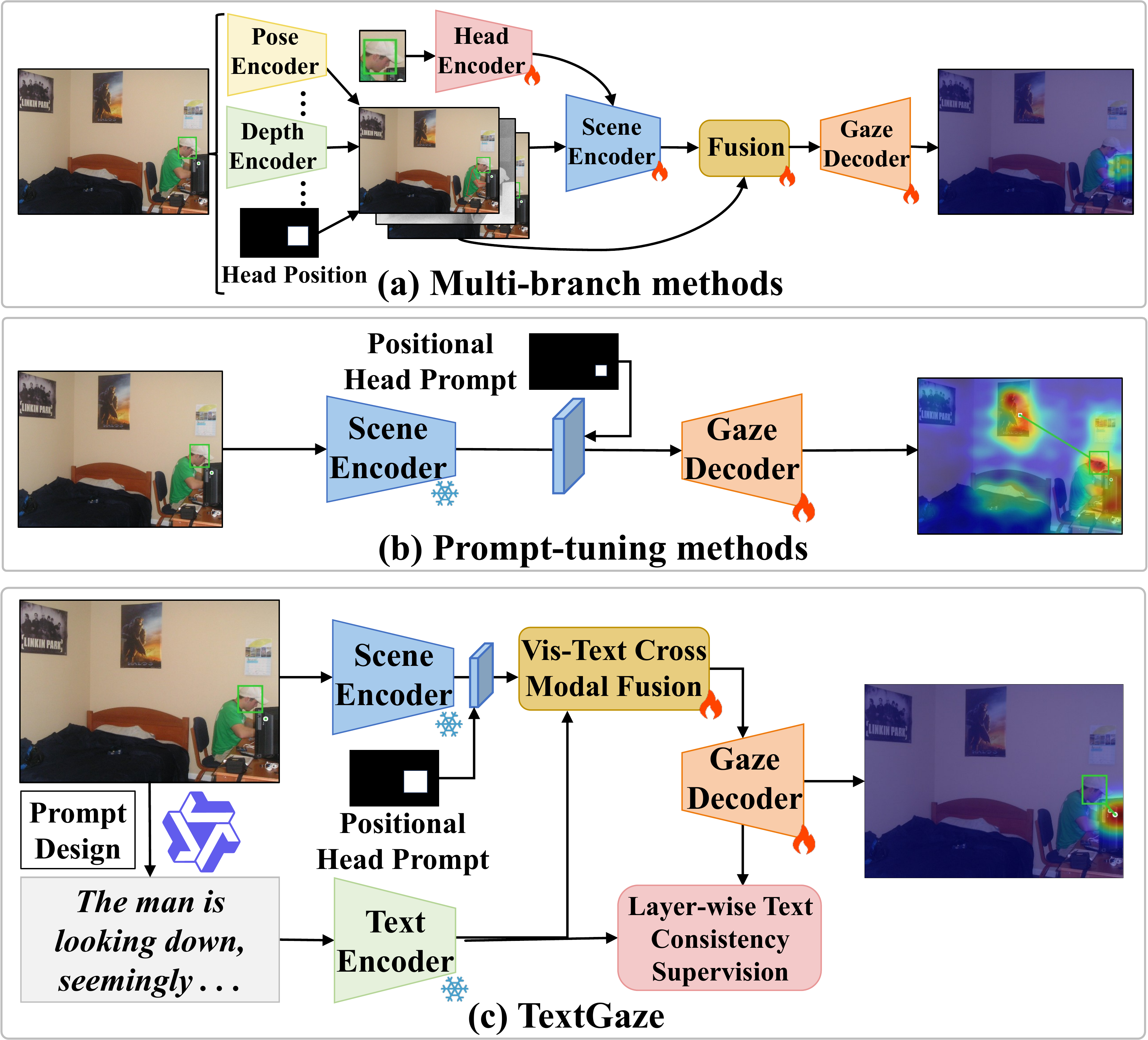}}
\caption{
Pipelines for gaze target estimation.
(a) Multi-branch methods~\cite{chen2021gaze} suffer from~\textit{slow convergence} due to increased structural complexity.
(b) Prompt-tuning method~\cite{ryan2025gaze} encounter~\textit{semantic defocus} as they struggle to isolate task-relevant saliency via positional prompts. By contrast, our method (c) leverages accessible scene-level text to achieve precise semantic focus on the gaze target.
}
\label{fig:intro}
\end{figure}

In multi-branch methods, the classical architecture is the dual-stream appearance-based design~\cite{chong2018connecting,lian2018believe,chong2020detecting,wang2024eulermormer}. In these designs, the head stream localizes the subject, and the scene stream encodes contextual visual information. Because purely visual information is insufficient to enhance effects, this paradigm has evolved into multi-branch frameworks that incorporate auxiliary cues.  These include head features~\cite{chen2021gaze}, scene context~\cite{saran2018human}, depth~\cite{bao2022escnet}, and pose~\cite{gupta2022modular}. 

Although these designs enrich visual representations, they introduced the slow convergence problem~\cite{ryan2025gaze} as adding an extra branch requires additional task-specific supervision beyond gaze annotations.  As the number of branches increases, it not only increases the annotation burden and subjectivity but also complicates deployment and hinders domain transfer. More fundamentally, these systems function as post-processing geometric filters. They are capable of eliminating illogical regions but cannot reason about which objects constitute valid gaze targets. Consequently, the problem of semantic under-specification remains unresolved.

To simplify such complex architectures, recent work has introduced prompt-tuning methods based on foundation models. Gazelle~\cite{ryan2025gaze} replaces person-specific scene re-encoding and handcrafted fusion with a unified decoder conditioned on a head position prompt, thereby addressing the need to integrate head information with global scene cues while avoiding multi-branch designs. This strategy substantially reduces architectural complexity and improves convergence behavior. 
However, the performance of such methods is heavily influenced by the base scene encoder and remains constrained by geometric conditions. Head position prompts also fail to provide effective information regarding task semantics. 
Therefore, the reasoning process is still largely influenced by low-level appearance heuristics (such as spatial proximity and visual salience), leading to semantic defocus in Figure~\ref{fig:intro}, where predicted attention contradicts the subject’s actual gaze intention. 

Gaze following is inherently an intent-level reasoning problem rather than a purely geometric regression task. 
Cognitive studies~\cite{mayrand2024gaze} show that human gaze integrates spatial cues with semantic judgments of meaningful targets and mental-state reasoning about target visibility. Yet existing models emphasize geometric signals and largely neglect task-level semantics, highlighting the need for accessible semantic priors that can guide gaze inference without reintroducing multi-branch complexity.

With the rise of large language models and LVLMs~\cite{bai2025qwen2,wang2024emu3}, the capabilities of LVLMs are being recognized across various fields. LVLMs offer robust zero-shot capabilities for extracting contextual visual cues~\cite{gupta2024exploring} and alleviate vision-dominated semantic underspecification by framing gaze estimation as a top-down, semantics-guided attention task. Compared to specialized, multi-branch architectures, LVLMs offer a more scalable and streamlined approach to acquiring auxiliary visual information. Their text tokens encode abstract concepts, such as objects and actions, serving as differentiable, high-level priors~\cite{bi2025unveiling}. These priors are guided by carefully designed visual prompts that specify task objectives. They constrain the solution space, improve adaptation to complex scenes, and eliminate the need for redundant branches and extra annotations. This mitigates the imbalance in the multi-branch semantic compensation cost.

Building on these insights, we introduce TextGaze, a unified trainable decoder that integrates LVLMs, scene encoding, and visual-text fusion. 
The model adopts a dual-stream design, shifting gaze estimation from direct geometric regression (pixels to gaze coordinates) to an intent-recognition process via a semantic layer (pixels to semantic intent to gaze target). 
One stream uses an LVLM with a BERT encoder to generate textual descriptions of subject behavior and plausible targets, yielding semantically enriched visual-text embeddings; the other extracts fine-grained scene cues via a frozen scene encoder and utilizes positional head prompting to enhance character positioning and facial cues. 
These complementary streams are fused by a Visual-Text Fusion Transformer for context-aware, cross-modal gaze estimation. 
To enhance model capability and robustness, we devise multiple lightweight and effective optimization strategies.
Based on our design implementations, TextGaze strikes a balance between multi-branch methods and streamlined designs.
And TextGaze achieves the best performance on the majority of evaluation metrics, validating the effectiveness of our fusion strategy. 

In summary, the main contributions of this paper are:
\begin{itemize}
\item We introduce textual scene cues into gaze target estimation, providing auxiliary information to mitigate the issue of semantic defocus in the gaze decoder. 
\item We strengthen the GazeFollow and VideoAttentionTarget benchmarks by integrating scene-target-person attention fields with LVLM-oriented prompts; this approach generates robust contextual cues that effectively improve the accuracy of text-aided gaze estimation. 
\item We design a unified dual-stream cross-modal fusion framework that combines textual embeddings with fine-grained visual representations, improving spatial reasoning, scene adaptability, and mimicking human top-down attention.
\end{itemize}


\section{Related Work}
\subsection{Gaze Target Estimation}
Gaze target estimation~\cite{recasens2015they,recasens2017following,lin2025gazehta,liu2024depth,ryan2025gaze} aims to predict the location a person is looking at in a scene image~\cite{recasens2015they} or video frame~\cite{recasens2017following}. 
The straightforward pipeline is the dual-stream approach, where one stream focuses on scene understanding and the other learns head features from cropped regions via off-the-shelf detectors~\cite{lin2025gazehta}. This design addresses the core requirements of locating individuals and learning visual scene features, but falls short of outstanding performance due to its sole reliance on visual appearance.
Subsequently, multi-branch architectures are proposed to capture diverse auxiliary cues (\eg, facial or head features~\cite{chen2021gaze}, scene context~\cite{saran2018human}, depth structures~\cite{bao2022escnet}, pose information~\cite{gupta2022modular}, and eye~\cite{fang2021dual}), with multi-modal fusion becoming key for performance gains. 
However, complex multi-branch designs suffer from diminishing returns and high data acquisition costs, driving efforts to streamline architectures. 
Recently, the prompt-tuning method Gaze-LLE~\cite{ryan2025gaze} proposed to leverage visual foundation models for gaze prediction via positional head prompt. However, it still faces challenges in explicit visual modeling and task-relevant semantic focus. 
To address these issues, we propose TextGaze, a framework that leverages accessible, interpretable auxiliary semantic cues, enabling high-performance gaze target estimation through effective cross-modal fusion.

\subsection{Large Vision Language Models}
Large Vision Language Models (LVLMs)~\cite{radford2021learning,li2023blip,yang2023dawn} have become a central paradigm for multimodal visual
understanding~\cite{li2023vigt,li2023transformer}. Models such as CLIP~\cite{radford2021learning}, Claude~\cite{priyanshu2024ai}, and GPT-4V~\cite{yang2023dawn} exhibit strong multimodal reasoning capabilities and achieve competitive or superior performance to classical vision models in zero-shot classification. Their architectures increasingly rely on pre-trained large language models to align visual and textual representations~\cite{grattafiori2024llama}, supported by components such as vision encoders, text encoders, decoders, and cross-attention modules. 
Recent advances further strengthen cross-modal fusion by unifying modalities into tokenized streams~\cite{wang2024emu3} or employing transfusive mechanisms~\cite{zhou2024transfusion}. 
In our work, LVLMs are used not as task solvers but as generators of structured visual auxiliary information for gaze target estimation. 
The LVLM-derived descriptions are embedded together with images into a joint representation space. A fused transformer with hierarchical fine-grained supervision is then applied to integrate these signals. This design significantly enhances cross-modal fusion and yields strong performance on gaze target estimation benchmarks.

\begin{figure*}[t!]   
\centering   
\includegraphics[width=1\textwidth, keepaspectratio]{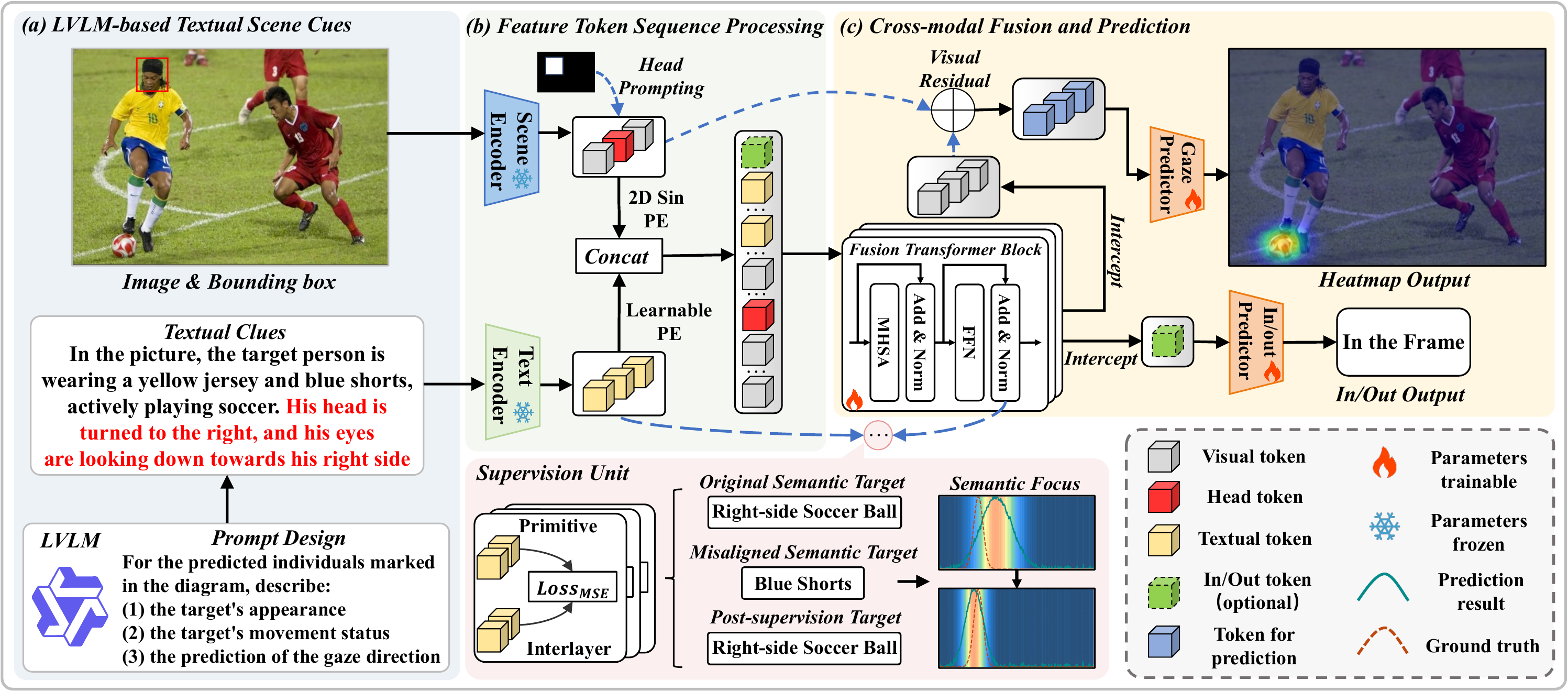} 
\caption{Overview of the proposed TextGaze. The diagram illustrates the end-to-end pipeline, including visual feature extraction via frozen DINOv2, text feature processing with pre-trained BERT, cross-modal fusion through a 3-layer fusion transformer, and final prediction heads for gaze heatmap and in/out status.}
\label{fig:architecture_overview}  
\end{figure*}


\section{Methodology}
\label{sec:method}
\subsection{Problem Formulation and Notation}
Gaze target estimation~\cite{recasens2017following,ryan2025gaze} aims to predict a heatmap that indicates the per-pixel probability of the gaze target. 
Formally, given an RGB image $\mI_{img} \in \sR^{H_{in} \times W_{in}\times 3}$
and the bounding box of the target person's head $\vx_{bbox}$, represented by length-4 coordinates,
the model predicts a 2D gaze heatmap $\mI_{heat} \in [0, 1]^{H \times W}$. 
If an in-/out-of-frame estimation task (\eg, VideoAttentionTarget and ChildPlay benchmarks) is included, the model should predict a scalar probability $p_y \in [0, 1]$. 
\subsection{Model Architecture}
\label{sec:architecture}
An architecture overview of the proposed method is shown in Figure~\ref{fig:architecture_overview}. 
It consists of three core components:
\textbf{Textual Scene Cues (\S\ref{sec:cues})}, which extracts attention-aware textual cues from a frozen LVLM.
\textbf{Feature token sequence processing (\S\ref{sec:token})}, responsible for enhancing visual features, embedding positional information, and constructing token sequences for fusion.
\textbf{Cross-modal fusion and prediction (\S\ref{sec:predict})}, which integrates the visual and textual sequences through fusion transformer blocks and produces a heatmap and in/out predictions. Below, we detail each component.

\subsection{LVLM-based Textual Scene Cues Module}
\label{sec:cues}
Obtaining semantically rich textual cues that align with visual focus is a key component of our design. We incorporate a frozen LVLM $\mathcal{L}$ into our design to accomplish this task. Specifically, we adopt Qwen2.5-VL-7B-Instruct in our experiments, while the framework is backbone-agnostic and compatible with any multimodal model supporting joint image-text prompting. Leveraging the zero-shot contextual reasoning and visual grounding ability of modern LVLMs, we design task-specific prompts to guide $\mathcal{L}$ in producing structured descriptions of the subject's appearance, behavior, and coarse gaze orientation. Instead of directly localizing gaze targets, which is unreliable in out-of-frame cases, this design generates context-aware auxiliary cues that remain effective across diverse scenarios.

The natural language outputs of $\mathcal{L}$ are encoded by a frozen text encoder, BERT-base-uncased in our implementation, into dense embeddings
$
\mS_t \in \sR^{T_t \times d_t},
$
where $T_t$ denotes the sequence length and $d_t$ the embedding dimension. These embeddings bridge high-level linguistic context and low-level visual features, enabling effective cross-modal fusion. Overall, the prompt design and LVLM-based cues module provide a flexible and robust solution for gaze target estimation in complex settings.

\subsection{Feature Token Sequence Processing Module}
\label{sec:token}
This module prepares text and visual features for bimodal fusion. For text processing, we augment the text feature sequence $\mF_t\in\sR^{L_t\times d}$ with learnable positional embeddings $\mathbf{PE}_{text}\in\sR^{d}$, obtaining $\mZ_t = \mF_t + \mathbf{PE}_{text}$.
For visual feature processing, we use a frozen DINOv2 encoder to extract a visual feature map $\mF_v\in\sR^{H\times W\times d}$. To enhance sensitivity to the target person's head, we incorporate head-region semantic cues using a learnable head token embedding. 
Given the head bounding box, we generate a downsampled binary mask $\mM \in \{0,1\}^{H \times W}$ for the head region. We then add a learnable head embedding $\mathbf{E}_{head}$ to the visual features via masked modulation: $\mZ_v = \mF_v + \mM \cdot \mathbf{E}_{head}$, where $\cdot$ denotes broadcasted element-wise multiplication.
The resulting feature map is flattened into a sequence $\mS_v$, and absolute 2D sinusoidal positional embeddings $\mathbf{PE}_v$ (fixed for the visual feature size) are added to form $\mS_{ve} = \mS_v + \mathbf{PE}_v$.

For fusion, we concatenate the text and visual sequences to construct the bimodal fusion sequence
$
\mS_f = \mathrm{Concat}\!\left(\mS_{te}, \mS_{ve}\right)
$.
The concatenation is operated along the token dimension. 
We record index boundaries to extract the visual segment for subsequent prediction. For the in-/out-of-frame classification task, a learnable token $\vz_{in/out}\in \sR^{d}$ is prepended, yielding $\mZ_{joint}\in\sR^{(1+HW+L_t)\times d}$:
\begin{equation}
\mZ_{joint} = [\vz_{in/out}, \underbrace{\vz_1, \vz_2, \cdots, \vz_{L_t}}_{\mZ_t}, \underbrace{\vz_1, \vz_2, \cdots, \vz_{HW}}_{\mZ_v}].
\end{equation}

\subsection{Cross-Modal Fusion and Prediction Module}
\label{sec:predict}
During the design of cross-modal fusion modules, we observed inherent challenges in vision-text fusion based on transformers. Such models lack explicit hierarchical constraints, leading to an excessive focus on global token interactions, resulting in representation drift across layers~\cite{clark2019does,raganato2018analysis}. In vision-language settings, this manifests as token-level misalignment between modalities~\cite{lu2019vilbert}. During deep fusion, gaze-related textual semantics are gradually diluted. In the absence of intermediate supervision, this discrepancy arising from visual gaze cues and the a priori textual intent accumulates layer by layer, ultimately leading to misalignment.

To address these issues, we design a three-layer stacked fusion transformer with post-normalization. It is tailored for unified attention construction, global cross-modal interaction, and hierarchical text semantic supervision, enabling precise vision-text alignment while retaining task-critical text semantics. 

The query $\mQ$ and key $\mK$ are derived by adding positional embeddings to the input sequence, while the value $\mV$ retains the original features:
\begin{equation}
\mQ = \mK = \mX^l + \mathbf{E}_{\text{pos}}, \quad \mV = \mX^l .
\end{equation}
Multi-head self-attention (MHSA) follows the standard scaled dot-product formulation. Only padding masks are applied to remove invalid tokens, allowing unrestricted interaction between text and visual tokens.

\begin{table*}[t!]
\centering
\tabcolsep 8pt
\resizebox{1.0\linewidth}{!}{
\begin{tabular}{|l|c|ccc|ccc|}
\hline\thickhline
\rowcolor{gray!25} & & \multicolumn{3}{c|}{GazeFollow} & \multicolumn{3}{c|}{VideoAttentionTarget} \\
\rowcolor{gray!25} \multirow{-2}{*}{Method} & \multirow{-2}{*}{Input} & AUC $\uparrow$ & Avg L2 $\downarrow$ & Min L2 $\downarrow$ & AUC $\uparrow$ & L2 $\downarrow$ & AP$_{in/out}$ $\uparrow$ \\
\hline
\textit{One Human} & - & \textit{0.924} & \textit{0.096} & \textit{0.040} & \textit{0.921} & \textit{0.051} & \textit{0.925} \\ \hline
Recasens \etal~\cite{recasens2017following} & I & 0.878 & 0.190 & 0.113 & - & - & - \\
Chong \etal~\cite{chong2018connecting} & I & 0.896 & 0.187 & 0.112 & 0.833 & 0.171 & 0.712 \\
Lian \etal~\cite{lian2018believe} & I & 0.906 & 0.145 & 0.081 & - & - & - \\
Chong \etal~\cite{chong2020detecting} & I & 0.921 & 0.137 & 0.077 & 0.860 & 0.134 & 0.853 \\
Chen \etal~\cite{chen2021gaze} & I & 0.908 & 0.136 & 0.074 & - & - & - \\
Fang \etal~\cite{fang2021dual} & I+D+E & 0.922 & 0.124 & 0.067 & 0.905 & 0.108 & 0.896 \\
Bao \etal~\cite{bao2022escnet} & I+D+P & 0.928 & 0.122 & - & 0.885 & 0.120 & 0.869 \\
Jin \etal~\cite{jin2022depth} & I+D+P & 0.920 & 0.118 & 0.063 & 0.900 & 0.104 & 0.895 \\
Tonini \etal~\cite{tonini2022multimodal} & I+D & 0.927 & 0.141 & - & 0.862 & 0.125 & 0.742 \\
Hu \etal~\cite{hu2022gaze} & I+D+O & 0.923 & 0.128 & 0.069 & 0.880 & 0.118 & 0.881 \\
Gupta \etal~\cite{gupta2022modular} & I+D+P & 0.943 & 0.114 & 0.056 & 0.914 & 0.110 & 0.879 \\
Horanyi \etal~\cite{horanyi2023they} & I+D & 0.896 & 0.196 & 0.127 & 0.832 & 0.199 & 0.800 \\
Miao \etal~\cite{miao2023patch} & I+D & 0.934 & 0.123 & 0.065 & 0.917 & 0.109 & \underline{0.908} \\
Liu~\etal~\cite{liu2024depth} & I+D & 0.936 & 0.121 & 0.061 & 0.918 & 0.108 & 0.882 \\
Tafasca \etal~\cite{tafasca2023childplay} & I+D & 0.939 & 0.122 & 0.062 & 0.914 & 0.109 & 0.834 \\
Tafasca \etal~\cite{tafasca2024sharingan} & I & 0.944 & 0.113 & 0.057 & - & 0.107 & 0.891 \\
Ryan \etal (DINOv2 ViT-B)$^{\ast}$~\cite{ryan2025gaze} & I & 0.955 & 0.106 & 0.048 & 0.928 & 0.110 & 0.877 \\
Ryan \etal (DINOv2 ViT-L)$^{\ast}$~\cite{ryan2025gaze} & I & \textbf{0.957} & \underline{0.101} & \textbf{0.043} & \underline{0.937} & \underline{0.103} & 0.903 \\
\hline
\rowcolor{gray!10} \textbf{Ours (DINOv2 ViT-B)} & I+T & 0.953 & 0.105 & \underline{0.047} & 0.932 & 0.113 & 0.879 \\
\rowcolor{gray!10} \textbf{Ours (DINOv2 ViT-L)} & I+T & \underline{0.956} & \textbf{0.099} & \textbf{0.043} & \textbf{0.941} & \textbf{0.102} & \textbf{0.911} \\
\hline
\end{tabular}
}
\caption{Gaze target estimation results on GazeFollow and VideoAttentionTarget. Inputs: I (image), D (depth), E (eyes), P (pose), O (objects), and T (text). The best results are in \textbf{bold} and the second-best are \underline{underlined}. $^{\ast}$Results reproduced by us from the official code releases.}
\label{tab:sota}
\end{table*}

Each fusion block contains two sublayers: MHSA and a feed-forward network (FFN), both equipped with residual connections and layer normalization. The MHSA sublayer output is:
\begin{equation}
\mX'^l = \text{LN}\!\left(\mX^l + \text{Dropout}\!\left(\text{MHSA}(\mQ, \mK, \mV)\right)\right),
\end{equation}
followed by the FFN sublayer:
\begin{equation}
\mX^{l+1} = \text{LN}\!\left(\mX'^l + \text{Dropout}\!\left(\text{FFN}(\mX'^l)\right)\right).
\end{equation}

To safeguard task-relevant textual semantics from degradation during cross-modal fusion, we propose a hierarchical text feature consistency supervision mechanism that enforces semantic alignment across all transformer layers. For each layer $l \in \{1,2,\dots,L\}$, the text segment of $\mX^{l}$ is extracted and aligned with pre-computed text encoder features.

Let $\mX^{l}_{text} \in \sR^{L_{t} \times d}$ denote the text token subset, where $L_{t}$ is the number of text tokens. After permuting to $L_{t} \times d$, it is projected to the text hidden dimension $d_{text}$ via a layer-specific linear projection (weight matrix $\mW_l \in \sR^{d_{text} \times d}$, bias term $\mathbf{b}_l \in \sR^{d_{text}}$):
\begin{equation}
\hat{\mS}_t^l = \mX^{l}_{text} \mW_l^\top + \mathbf{b}_l ,
\end{equation}
where $\hat{\mS}_t^l \in \sR^{ L_{t} \times d_{text}}$ are the reconstructed text features.

The layer-wise text supervision loss is the mean squared error between $\hat{\mS}_t^l$ and target text features $\mS_t$:
\begin{equation}
\text{MSE}\!\left(\hat{\mS}_t^l, \mS_t\right)
= \frac{1}{N} \sum_{i=1}^N \left(\hat{\mS}_t^l[i] - \mS_t[i]\right)^2 ,
\end{equation}
where $N = L_{t} \cdot d_{text}$ is the total number of elements after vectorizing the text features. The hierarchical text loss is:
\begin{equation}
\mathcal{L}_{text} = \frac{1}{L} \sum_{l=1}^L \mathcal{L}_{text}^l .
\end{equation}

Following the fusion module, the visual token segment is reshaped into $\mF \in \mathbb{R}^{H \times W\times d}$, where $H$ and $W$ denote spatial dimensions. A two-layer convolutional decoder outputs the gaze probability heatmap $\mI_{heat}$, supervised by pixel-wise BCE loss with Gaussian targets ($\sigma{=}3$). For in-/out-of-frame classification, the task token $\vz_{in/out}$ is fed to a two-layer MLP $D_{in/out}$ to predict a Bernoulli probability.

The overall objective is:
\begin{equation}
\mathcal{L}_{\text{all}} = \mathcal{L}_{\text{heat}} + \lambda_1 \mathcal{L}_{in/out} + \lambda_2  \mathcal{L}_{text},
\end{equation}
where $\lambda_1$ and $\lambda_2$ balance the auxiliary losses.


\section{Experiments}
\subsection{Experimental Setup}
\textbf{Datasets.} We evaluate our method on four benchmark datasets: GazeFollow~\cite{recasens2015they}, VideoAttentionTarget~\cite{chong2020detecting}, ChildPlay~\cite{tafasca2023childplay}, and GOO-Real~\cite{tomas2021goo}.
GazeFollow contains 117,727 training images with 124,095 annotated target persons and 4,782 test images. VideoAttentionTarget consists of 58,507 training frames with 132,563 target persons and 13,127 test frames with 31,978 target persons. ChildPlay and GOO-Real are used to assess generalization without fine-tuning, focusing on children’s gaze behaviors and everyday shopping scenarios, respectively. 
To construct visual descriptive cues, we employ prompt engineering to guide Qwen2.5-VL-7B~\cite{bai2025qwen2} to generate appearance attributes for each image. Specifically, these attributes include (1) appearance descriptions of the target person for multi-person disambiguation; (2) action descriptions to better capture gaze behavior; and (3) gaze direction estimates without naming the target object, avoiding hallucination when the target is out of frame. 
Detailed implementation is given in the appendix of the supplementary material.

\noindent\textbf{Evaluation Protocols.} 
Following existing gaze target estimation settings~\cite{chong2018connecting,tonini2022multimodal,miao2023patch,ryan2025gaze}, we use heatmap AUC and pixel-wise L2 distance as the primary metrics.
For heatmap AUC, each pixel in the predicted heatmap is treated as a confidence score to compute the ROC curve. The pixel-wise L2 distance is defined as the Euclidean distance between the argmax location of the predicted heatmap and the ground-truth gaze target. 
For the GazeFollow dataset, as it has multiple annotation points, we compute the distance to the mean of annotations (Avg L2) and the distance to the nearest annotation (Min L2). 
For the VideoAttentionTarget and ChildPlay datasets, we also evaluate the Average Precision (AP) for the in-/out-of-frame prediction task to assess whether the gaze target lies within the visual field.

\noindent\textbf{Implementation Details.} The input image is resized to 448$\times$448, while the predicted gaze heatmap has a resolution of 64$\times$64. 
The scene encoder uses frozen DINOv2~\cite{oquab2023DINOv2} optimized by Adam~\cite{kingma2015adam} optimizer with the initial learning rate of 1e-4 and weight decay of 1e-2, including ViT-B and ViT-L versions. The text encoder utilizes a frozen BERT-base-uncased model. Default loss weights are $\lambda_1 = 1$ and $\lambda_2=0.1$.

\begin{figure*}[t!]
\begin{center}
\centerline{\includegraphics[width=1\linewidth]{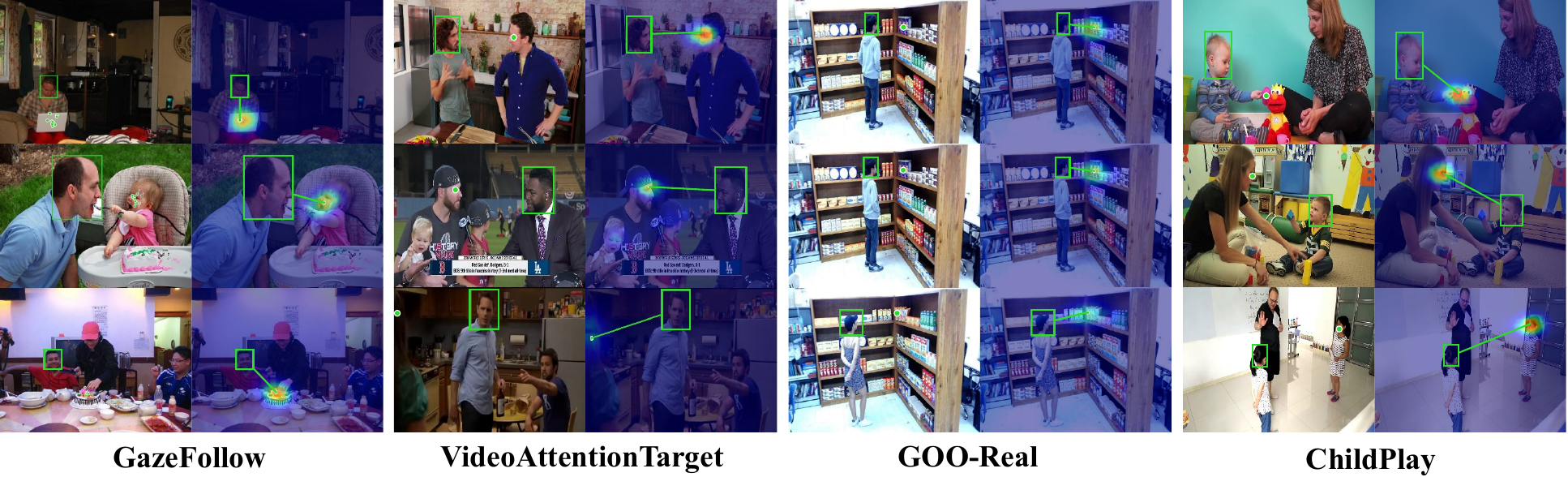}}
\caption{
Qualitative results on the GazeFollow, VideoAttention-Target (fine-tuned), and ChildPlay/GOO-Real (without fine-tuning) datasets. The input image on the left and the predicted heatmaps on the right. Our method shows consistent accuracy on different scenes. 
}
\label{fig:vis_results}
\vspace{-1.5em}
\end{center}
\end{figure*}

\subsection{Performance Comparison}
\textbf{Comparison to State-of-the-Art.} 
Our method's performance on the GazeFollow and VideoAttentionTarget datasets compared with existing leading approaches is summarized in Table~\ref{tab:sota}. It is easy to notice that our model differs from most prior methods, which rely on additional modalities such as depth or pose; instead, we take image and text as inputs. This design originates from our thinking on balancing the acquisition of auxiliary cues and architectural complexity.

On GazeFollow, our method achieves competitive results across all metrics. With the DINOv2 ViT-L backbone, we obtain the lowest average $\mathcal{L}_2$ error and match the best minimum $\mathcal{L}_2$, while delivering a high AUC that is comparable to the state-of-the-art. Even with the lighter ViT-B backbone, our model maintains top-tier performance on key metrics. These results validate our hypothesis that the text-guided semantic priors provided by LVLMs effectively supply auxiliary cues, thereby enhancing the performance of gaze target estimation models. 
On VideoAttentionTarget, our DINOv2 ViT-L variant outperforms existing approaches across all three evaluation metrics (AUC, $\mathcal{L}_2$ error, and AP$_{\text{in/out}}$), demonstrating the value of integrating text cues for reliable gaze estimation. Our method particularly excels in the in-/out-of-frame task, highlighting its ability to determine whether a target lies within the field of view. This conclusion is further validated by cross-dataset comparison results.

It should be noted that fair comparisons with the state-of-the-art method by Ryan~\etal~\cite{ryan2025gaze} are conducted under matched experimental settings using DINOv2 ViT-B and DINOv2 ViT-L backbones, with their results reproduced from official code. Qualitative examples under the DINOv2 ViT-B configuration are provided in Section~\ref{sec:qualitative} to further illustrate our approach’s effectiveness.

\begin{table}[t!]
\centering
\tabcolsep 7pt
\resizebox{1.0\linewidth}{!}{%
\begin{tabular}{|l|cc|}
\hline\thickhline
\rowcolor{gray!25}
Method & AUC $\uparrow$ & L2 $\downarrow$ \\
\hline
Chong \etal~\cite{chong2020detecting} & 0.670 & 0.334 \\
Tonini \etal~\cite{tonini2022multimodal} & 0.840 & 0.238 \\
Miao \etal~\cite{miao2023patch} & 0.869 & 0.202 \\
Ryan \etal (DINOv2 ViT-B)~\cite{ryan2025gaze} & 0.901 & \underline{0.174} \\
Ryan \etal (DINOv2 ViT-L)~\cite{ryan2025gaze} & 0.898 & 0.175 \\
\hline
\rowcolor{gray!10} \textbf{Ours (DINOv2 ViT-B)}& \underline{0.906} & 0.179 \\
\rowcolor{gray!10} \textbf{Ours (DINOv2 ViT-L)} & \textbf{0.918} & \textbf{0.159} \\
\hline
\end{tabular}}
\caption{Cross-dataset results on the GOO-Real dataset. The best results are in \textbf{bold} and the second-best are underlined.}
\label{tab:goo_real_results}
\vspace{-1.5em}
\end{table}

\noindent\textbf{Cross-dataset Results.}
To evaluate the generalization capability of our approach, we conduct cross-dataset evaluation on the GOO-Real and ChildPlay datasets without modifying LVLM prompts or performing training, fine-tuning, or data augmentation.
\ding{182} The test scenarios in the GOO-Real dataset involve gaze targets within the field of view. Table~\ref{tab:goo_real_results} compares our results with existing methods, where our approach achieves state-of-the-art performance on both AUC and L2 metrics, demonstrating strong generalization in real-world scenarios.
\ding{183} The ChildPlay dataset includes predictions of gaze target presence within the visual field of view. Table~\ref{tab:childplay_combined_results} compares our results with existing methods: although our approach slightly lags behind the state of the art in AUC and L2, it outperforms all baselines on the AP metric. Combined with results from the VideoAttentionTarget, this further highlights the contribution of LVLM-derived visual descriptive cues to AP (in/out) classification. Figure~\ref{fig:vis_results} presents qualitative results across four datasets: GazeFollow (trained with DINOv2 ViT-B), VideoAttentionTarget (fine-tuned), and ChildPlay/GOO-Real (without fine-tuning). The left panel shows raw inputs, while the right overlays predicted heatmaps.

\begin{table}[t!]
\centering
\tabcolsep 8pt
\resizebox{1.0\linewidth}{!}{%
\begin{tabular}{|l|ccc|}
\hline\thickhline
\rowcolor{gray!25}
Method & AUC $\uparrow$ & L2 $\downarrow$ & AP $\uparrow$ \\
\hline
Chong \etal~\cite{chong2020detecting} & 0.912 & 0.121 & - \\
Miao \etal~\cite{miao2023patch} & 0.933 & 0.113 & - \\
Gupta \etal~\cite{gupta2024exploring} & 0.923 & 0.142 & 0.694 \\
Tafasca \etal~\cite{tafasca2024sharingan} & 0.932 & 0.115 & 0.600 \\
Ryan \etal (ViT-B)~\cite{ryan2025gaze} & 0.941 & 0.119 & 0.993 \\
Ryan \etal (ViT-L)~\cite{ryan2025gaze} & \textbf{0.951} & \textbf{0.103} & \underline{0.994} \\
\hline
\rowcolor{gray!10} \textbf{Ours (DINOv2 ViT-B)} & 0.945 & 0.112 & 0.992 \\
\rowcolor{gray!10} \textbf{Ours (DINOv2 ViT-L)} & \underline{0.950} & \underline{0.111} & \textbf{0.996} \\
\hline
\end{tabular}}
\caption{Cross-dataset results on the ChildPlay dataset. The best results are in \textbf{bold} and the second-best are underlined.}
\label{tab:childplay_combined_results}
\end{table}

\subsection{Ablation Study}
\begin{figure}[t!]
\begin{center}
\centerline{\includegraphics[width=\columnwidth]{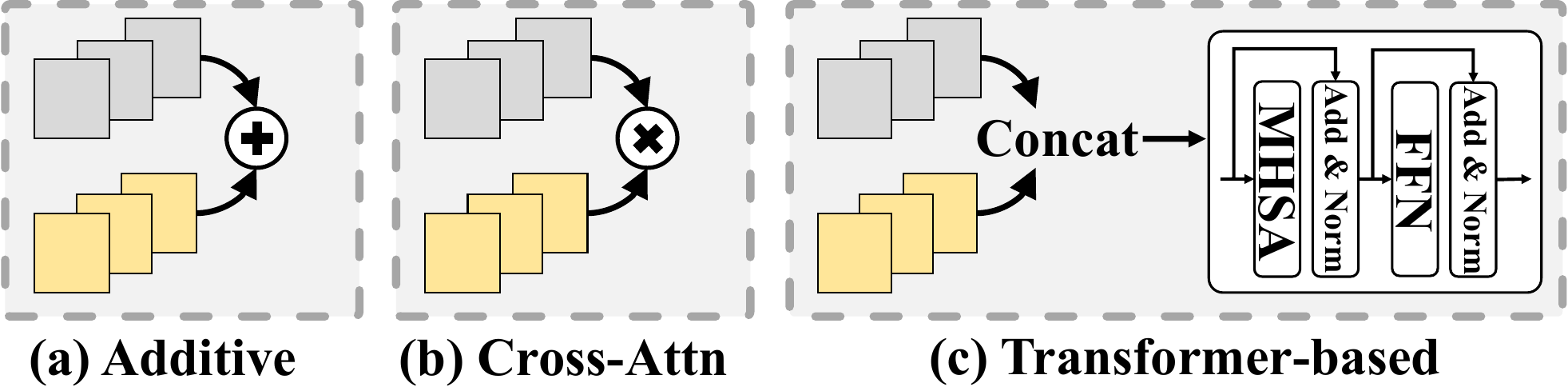}}
\caption{
Schematic of fusion strategy under ablation experiments. The grey and yellow blocks represent visual tokens and textual tokens, respectively.
}
\label{fig:fusion ablation}
\end{center}
\end{figure}

\textbf{Ablation Results of Bimodal Fusion.}
We explore multiple representative fusion approaches that align with our design principles to formulate our fusion strategy: additive fusion, cross-attention fusion, and the transformer-based fusion adopted in our model. Their schematic structures are illustrated in Fig.~\ref{fig:fusion ablation}. Additive fusion enforces element-wise alignment between modalities and thus struggles to handle heterogeneous visual-textual representations. By comparison, cross-attention mechanisms enable directional interactions, yet remain constrained by asymmetric information flow and lack holistic cross-modal reasoning capabilities. The transformer-based fusion performs unified attention across all visual and textual tokens, allows globally symmetric interactions, and progressively optimizes gaze-related semantic information. The relevant experimental results in Table~\ref{tab:ablation_study_f} demonstrate the effectiveness of this fusion strategy within our design. The ablation studies in this section are all conducted using the GazeFollow dataset. All tested models use the same DINOv2 ViT-B scene encoder and textual cues data.

\noindent\textbf{Ablation Results of Key Components.}
Building upon the established fusion strategy, we design key components to enhance models' performance in gaze target estimation tasks. 
We first focus on the critical semantic guidance capability of textual cues within tasks, noting that gaze-related textual cues may undergo semantic drift during multi-layer fusion processes. Consequently, we design an inter-layer text supervision mechanism as the foundational component and test its efficacy.
We also wish to achieve spatial fidelity preservation concisely. So, we introduce pre-prediction visual residual connections to preserve the fine-scale spatial details crucial for gaze localization.
Regarding positional embedding strategies for feature sequences, we attempt to incorporate explicit spatial inductive biases to model gaze-specific localization patterns. We test two approaches: modality-specific positional embedding and learnable full-sequence positional embedding. The former utilizes 2D sinusoidal positional embedding for visual features and learnable positional embedding for textual features; the latter applies learnable positional embedding to all features.
Each component we designed effectively enhances the overall performance of the architecture, with relevant results shown in Table~\ref{tab:ablation_study}. Stepwise optimization experiments further validate the rationality and effectiveness of our design approach. The results further indicate that simultaneously activating these three components yields optimal performance, confirming their complementary roles in our framework design for the gaze target estimation task. We ultimately adopted the simultaneous activation scheme for our architectural implementation.

\begin{table}[t!]
\centering
\tabcolsep 10pt  
\resizebox{1.0\linewidth}{!}{%
\begin{tabular}{|l|ccc|}  
\hline\thickhline
\rowcolor{gray!25}
Bimodal Fusion Strategy & AUC $\uparrow$ & Avg L2 $\downarrow$ & Min L2 $\downarrow$ \\
\hline
Additive Fusion   & 0.950 & 0.118 & 0.056 \\
Cross-Attention Fusion  & 0.952 & 0.119 & 0.055 \\
\hline
\rowcolor{gray!10} Transformer-based Fusion & \textbf{0.953} & \textbf{0.105} & \textbf{0.047} \\  
\hline
\end{tabular}}
\caption{Ablation study on bimodal fusion strategies. The best results are in \textbf{bold}.}  
\label{tab:ablation_study_f}
\end{table}

\begin{table}[t!]
\centering
\tabcolsep 6pt
\resizebox{1.0\linewidth}{!}{%
\begin{tabular}{|ccc|ccc|}
\hline\thickhline
\rowcolor{gray!25}
Txt Sup. & Vis Res. & Vis 2D PE & AUC $\uparrow$ & Avg L2 $\downarrow$ & Min L2 $\downarrow$ \\
\hline
$\times$ & $\times$ & $\times$ & 0.950 & 0.119 & 0.057 \\
$\checkmark$ & $\times$ & $\times$ & 0.951 & 0.117 & 0.057 \\
$\checkmark$ & $\checkmark$ & $\times$ & 0.952 & 0.113 & 0.053 \\
$\checkmark$ & $\times$ & $\checkmark$ & 0.952 & 0.112 & 0.052 \\
\hline
\rowcolor{gray!10} $\checkmark$ & $\checkmark$ & $\checkmark$ & \textbf{0.953} & \textbf{0.105} & \textbf{0.047} \\
\hline
\end{tabular}}
\caption{Ablation study on key components under transformer-based fusion strategy. Txt Sup., Vis 2D PE, Vis 2D PE denote text supervision, visual residual, and mixed positional embedding, respectively. The best results are in \textbf{bold}.}
\label{tab:ablation_study}
\end{table}

\subsection{Qualitative Results}
\label{sec:qualitative}
\begin{figure}[t!]
\begin{center}
\centerline{\includegraphics[width=\columnwidth]{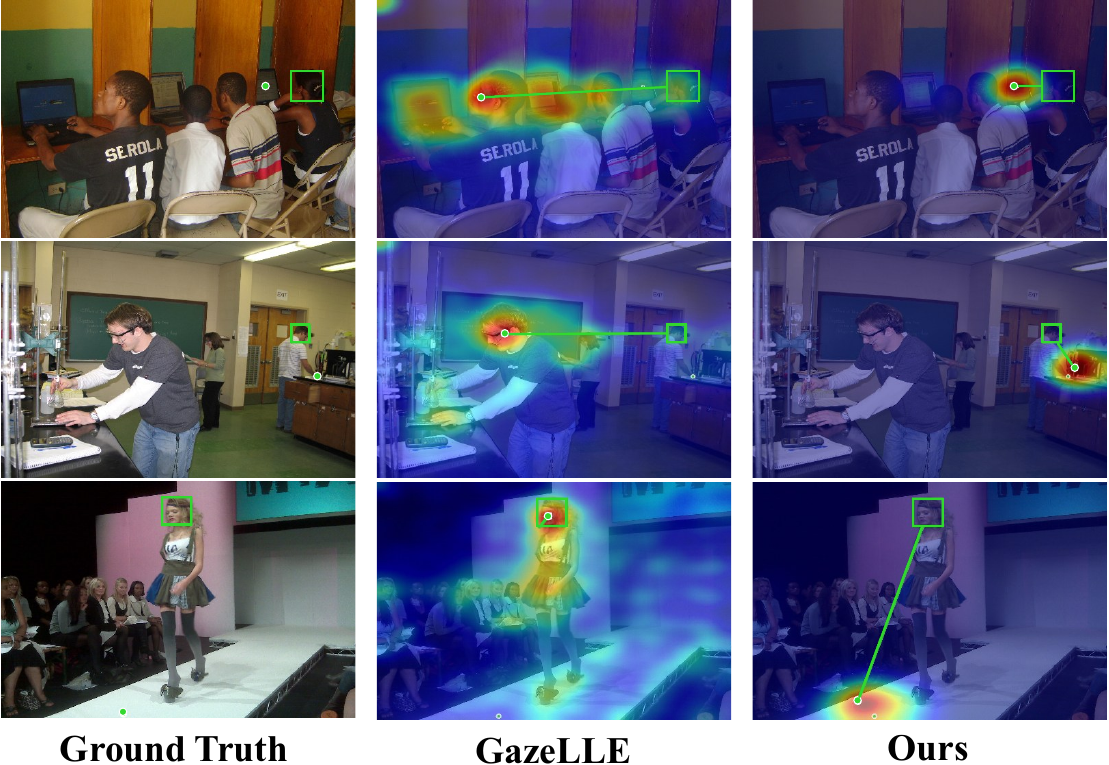}}
\caption{Qualitative comparison on the GazeFollow dataset. 
Compared to GazeLLE, our method demonstrates its superiority in three representative scenarios.
}
\label{fig:compare gazelle}
\end{center}
\end{figure}

Figure~\ref{fig:compare gazelle} presents three representative scenarios from GazeFollow that illustrate the qualitative advantages of TextGaze. In the scene depicted in the first row of the image, multiple targets and ambiguous candidate regions are present. Purely vision-guided methods are disrupted by visually salient areas due to the absence of explicit task-oriented semantics. By incorporating textual cues from LVLM to assign subject identity and gaze intent, our model is able to avoid misjudgments. In the second typical scenario, facial cues are rendered unreliable by occlusion or low resolution. Consequently, low-level visual cues are insufficient to support the precise predictions of purely vision-guided methods. However, our approach remains reliable through LVLM-based behavioural descriptions that compensate for missing details with high-level semantic priors. Finally, the target was misclassified as being within the bounding box, a frequent error arising from its reliance on spatial proximity. Our approach prevents such errors by integrating scene context with semantic constraints.
These cases validate the efficacy of our approach. Implementations that rely excessively on purely bottom-up visual understanding lack task-oriented semantics, resulting in semantic defocusing. By integrating prior knowledge guided by LVLM, we are able to shift gaze estimation from geometric regression towards a semantically driven decision process. This enables us to achieve accurate localization in complex scenes.

\section{Conclusion}
We address two issues in gaze target estimation: multi-branch designs require extensive supervision and annotations, causing cost imbalance and slow convergence; streamlined designs over-rely on low-level visual saliency rather than true gaze intent, leading to semantic defocus. We propose TextGaze, a unified cross-modal framework guided by an LVLM. Exploiting zero-shot capability, we produce text prompts and fuse them with frozen visual features through a dual-stream structure, enabling effective fusion while keeping the pipeline simple. We augment four datasets with text cues for future research. Experiments show strong performance and generalization, verifying our design and demonstrating LVLMs’ ability to connect vision with high-level intentions.

\section*{Acknowledgements}
This work was supported by the Key Science \& Technology Project of Anhui Province (202304a05020068), the Yangtze River Delta Science and Technology Innovation Community Joint Research (Basic Research) Project under Grant (2025CSJZN01600), and the New Generation of Artificial Intelligence National Science and Technology Major Project (2025ZD0123303).

\bibliographystyle{named}
\bibliography{ijcai26}

\end{document}